\documentclass{article}

\usepackage[final]{corl_2025} % Uncomment for the camera-ready ``final'' version.
\usepackage{amsmath,amsfonts,bm}

\def\eqref#1{equation~\ref{#1}}
\def\1{\bm{1}}

\DeclareMathAlphabet{\mathsfit}{\encodingdefault}{\sfdefault}{m}{sl}
\SetMathAlphabet{\mathsfit}{bold}{\encodingdefault}{\sfdefault}{bx}{n}

\usepackage{color,xcolor}
\usepackage{epsfig}
\usepackage{graphicx}

\usepackage{adjustbox}
\usepackage{array}
\usepackage{booktabs}
\usepackage{colortbl}
\usepackage{wrapfig}
\usepackage{hhline}
\usepackage{multirow}
\usepackage{amsmath,amsfonts,amssymb}
\usepackage{bm}
\usepackage{nicefrac}
\usepackage{microtype}
\usepackage{mathtools}

\usepackage{changepage}
\usepackage{extramarks}
\usepackage{fancyhdr}
\usepackage{lastpage}
\usepackage{setspace}
\usepackage{soul}
\usepackage{xspace}

\usepackage[pagebackref=true,breaklinks=true,colorlinks,citecolor=gray]{hyperref}

\usepackage{url}

\usepackage{enumerate}
\usepackage{enumitem}  % control indent of enumerate, itemize, etc.
\usepackage{makecell}

\usepackage{pifont} % for extra symbols such as \cmark

\usepackage{algorithm,algpseudocode}

\usepackage{amsthm}
\usepackage{float}

\usepackage[bottom]{footmisc}

\usepackage{listings}
\usepackage{xcolor}
\newcolumntype{L}[1]{>{\raggedright\let\newline\\\arraybackslash\hspace{0pt}}m{#1}}
\newcolumntype{C}[1]{>{\centering\let\newline\\\arraybackslash\hspace{0pt}}m{#1}}
\newcolumntype{R}[1]{>{\raggedleft\let\newline\\\arraybackslash\hspace{0pt}}m{#1}}

\newcommand{\ignore}[1]{}

\def\naive{na\"{\i}ve\xspace}

\DeclareMathAlphabet{\mathbfit}{OML}{cmm}{b}{it}

\makeatletter
\DeclareRobustCommand\onedot{\futurelet\@let@token\@onedot}
\def\@onedot{\ifx\@let@token.\else.\null\fi\xspace}

\def\eg{e.g\onedot} 
\def\ie{i.e\onedot} 
 
\def\etc{etc\onedot}

\makeatother

\definecolor{MyDarkBlue}{rgb}{0,0.08,1}
\definecolor{MyAqua}{rgb}{0,0.7,0.7}
\definecolor{MyDarkGreen}{rgb}{0.02,0.6,0.02}
\definecolor{MyDarkRed}{rgb}{0.8,0.02,0.02}
\definecolor{MyDarkOrange}{rgb}{0.40,0.2,0.02}
\definecolor{MyPurple}{RGB}{111,0,255}
\definecolor{MyRed}{rgb}{1.0,0.0,0.0}
\definecolor{MyGold}{rgb}{0.75,0.6,0.12}
\definecolor{MyDarkgray}{rgb}{0.66, 0.66, 0.66}
\definecolor{JiayuanColor}{rgb}{0.60,0.43,0.48}

\newcommand{\xhdr}[1]{{\noindent\bfseries #1}}

\newcommand{\squishlist}{
    \begin{list}
    { 
        \setlength{\itemsep}{0pt}
        \setlength{\parsep}{1pt}
        \setlength{\topsep}{1pt}
        \setlength{\partopsep}{0pt}
        \setlength{\leftmargin}{1em}
        \setlength{\labelwidth}{1em}
        \setlength{\labelsep}{0.5em} 
    }
}

\newcommand{\squishend}{
  \end{list}  
}

\definecolor{LightCyan}{rgb}{0.88,1,1}

\usepackage{soul}

\usepackage{pgfgantt}
\usepackage{MnSymbol}%
\usepackage{wasysym}%
\usepackage[capitalise]{cleveref}
\usepackage{caption}
\usepackage{subcaption}

\usepackage{dsfont}
\newcommand{\kw}[1]{{\textsc{\MakeLowercase{#1}}}}
\newcommand{\oursplain}{MultiGen\xspace}
\newcommand{\ours}{\kw{MultiGen}\xspace}

\newcommand{\diffp}{\kw{Diffusion Policy}\xspace}

\title{The Sound of Simulation: Learning Multimodal Sim-to-Real Robot Policies with Generative Audio}

\author{
Renhao Wang \And
Haoran Geng \And
Tingle Li \AND
Feishi Wang \And
Gopala Anumanchipalli \And
Trevor Darrell \AND
Boyi Li \And
Pieter Abbeel \And
Jitendra Malik \And
Alexei A. Efros
}

\affiliation{University of California, Berkeley}

\begin{document}
\maketitle

\begin{abstract}
    Robots must integrate multiple sensory modalities to act effectively in the real world. Yet, learning such multimodal policies at scale remains challenging. Simulation offers a viable solution, but while vision has benefited from high-fidelity simulators, other modalities (\eg sound) can be notoriously difficult to simulate. As a result, sim-to-real transfer has succeeded primarily in vision-based tasks, with multimodal transfer still largely unrealized. In this work, we tackle these challenges by introducing \ours, a framework that integrates large-scale generative models into traditional physics simulators, enabling multisensory simulation. We showcase our framework on the dynamic task of robot pouring, which inherently relies on multimodal feedback. By synthesizing realistic audio conditioned on simulation video, our method enables training on rich audiovisual trajectories---without any real robot data. We demonstrate effective zero-shot transfer to real-world pouring with novel containers and liquids, highlighting the potential of generative modeling to both simulate hard-to-model modalities and close the multimodal sim-to-real gap. 
    Code, models and data available at: \url{https://multigen-audio.github.io}
\end{abstract}

% Two or three meaningful keywords should be added here

\section{Introduction}
\label{sec:intro}
% \vspace{-0.5em}
Multimodal perception is essential for robust and adaptive human behavior, whether it be grasping an object by sight and feel, or pouring a drink while relying on auditory cues. For robots to achieve similar generalization and adaptability, they must also learn to integrate multiple sensory inputs effectively. However, acquiring large-scale multimodal datasets for robot learning is a significant challenge. Synchronized video, audio, tactile and action data require precise calibration, expensive hardware, and labor-intensive setup. As a result, even recent efforts to collect robot datasets at unprecedented scale still lack diverse sensory constellations~\cite{khazatsky2024droid, o2023open}.

A promising alternative to costly real-world data collection is simulation-based learning, where robots acquire skills in synthetic environments before adapting them to the real world via sim-to-real transfer. This strategy has enabled impressive progress in vision-based policy learning, particularly for tasks like locomotion and basic manipulation (\eg grasping or pick-and-place). However, more dexterous and dynamic behaviors---which often require multimodal feedback---remain out of reach. Sim-to-real transfer for such multimodal tasks has yet to be fully realized, due in large part to the difficulty of simulating non-visual modalities. Sound, for example, is notoriously hard to model in simulation: its physical propagation depends on complex wave dynamics and material interactions, making high-fidelity audio simulation both expensive and impractical at scale.

\begin{figure}[tp!]
\centering
\scalebox{1.0}{
    \includegraphics[width=\linewidth]{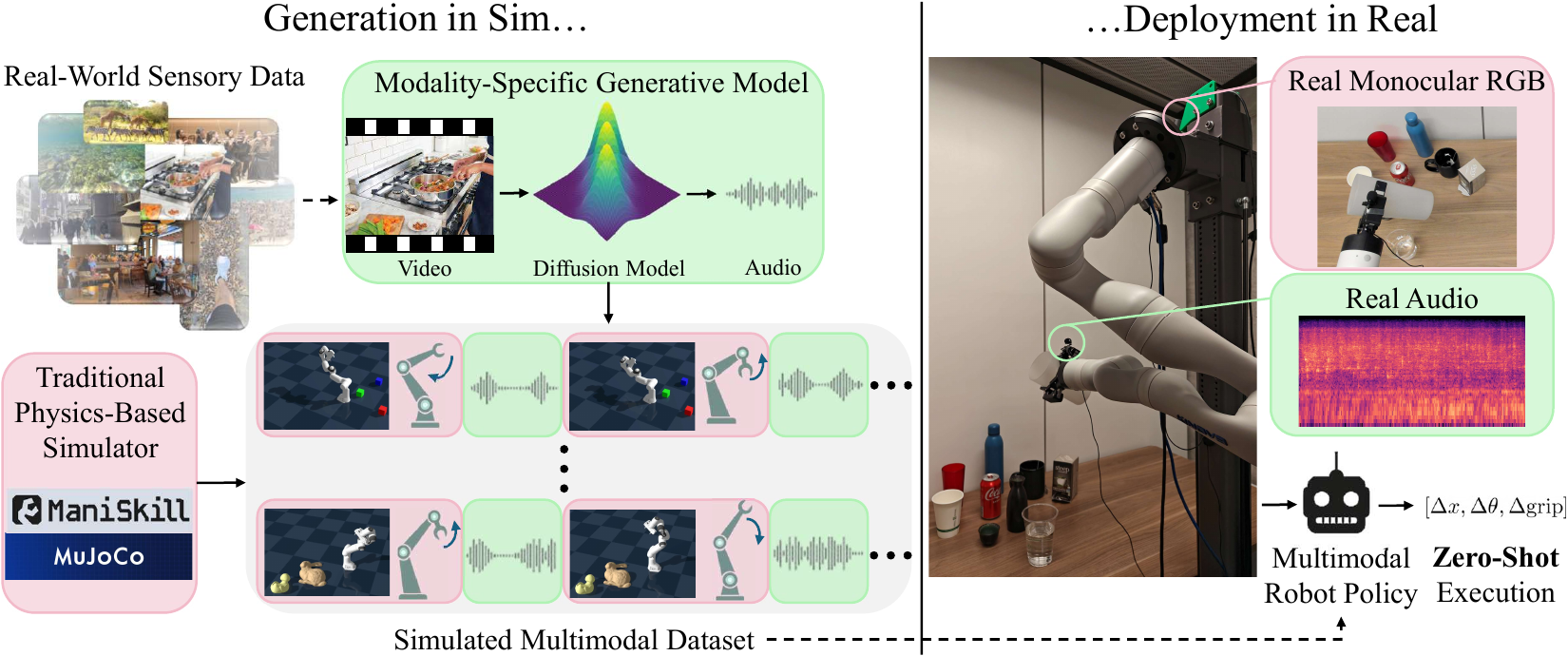}
}
\vspace{-1.5em}
\caption{\textbf{Overview of our \ours framework.} We train a generative model on real-world sensory data to capture modalities that would otherwise be difficult to simulate (\eg audio). Augmenting traditional simulators with these generative models enables generating synthetic multimodal data at scale, and learning multimodal policies that can translate \emph{zero-shot} to the real world.
}
% \vspace{-1em}
\vspace{-1.5em}
% \vspace{-0.1in}
\label{fig:teaser}
\end{figure}

In this work, we address both challenges---multimodal simulation and sim-to-real transfer---with \ours, a framework which augments traditional physics-based simulators with large-scale pretrained generative models. 
Specifically, \ours is a hybrid pipeline that runs a generative model in parallel with a physics engine, allowing for the synthesis of realistic, task-relevant sensory signals that complement visual and action data. To demonstrate the effectiveness of this approach, we instantiate \ours on the task of robot pouring, an inherently multimodal task that strongly relies on auditory feedback due to liquid occlusion, transparency, and visual ambiguity. Our generative model is pretrained on diverse real-world audiovisual data and finetuned on in-the-wild human pouring videos, enabling it to produce task-relevant sounds that enhance policy learning. We show that \ours enables high-quality audio synthesis that more faithfully matches real-world pouring sounds compared to standard data augmentation techniques. We also demonstrate that policies trained with \ours transfer effectively to real-world pouring of novel liquid types across diverse container geometries in a zero-shot manner---without requiring any real-world robot sensory or action data. These results highlight the potential of \ours to close the multimodal data gap, paving the way for learning more perceptually rich robotic systems at scale.

\vspace{-0.5em}
\section{Related Work}
\label{sec:related}
\vspace{-0.5em}
\xhdr{Multimodal robot learning.}
There has been increasing interest in integrating multiple sensory modalities for robotics.
Prior work has heavily explored leveraging vision and touch across various tasks, such as grasping objects~\cite{calandra2018more, calandra2017feeling}, representation learning~\cite{yang2024binding}, or for fine motor control~\cite{lee2019making, lin2024learning, qi2023general, tian2019manipulation}. But audio sensing has received relatively little attention in robotics. Most prior works incorporating sound focus on speech-based interaction~\cite{thomason2020jointly, shi2024yell} or environment and action recognition~\cite{gandhi2020swoosh, gan2020look}. Few studies leverage audio directly for manipulation, such as using impact sounds for material classification \cite{clarke2022diffimpact, clarke2018learning} or pouring sounds to estimate liquid quantity~\cite{liang2019making}. 
More recently, interest in contact-rich tasks with heavy occlusion and clear acoustic signal, such as dynamic pouring or object search and retrieval from confined spaces, have brought policies employing vision, tactile and audio to the fore~\cite{li2022see, du2022play, liu2024maniwav}. While these approaches all use teleoperated datasets of limited size, our work in contrast leverages generative models for synthesizing realistic multimodal data at scale.

\xhdr{Simulation of sound.}
Traditional approaches to simulating sound in robotics and graphics rely on physics-based methods, such as numerical wave propagation~\cite{tsuchiya2007numerical, rungta2016syncopation} or ray tracing~\cite{schissler2016interactive}. While these methods can achieve high-fidelity audio, they are computationally expensive and often require detailed material properties~\cite{chen2022soundspaces, matl2020stressd}, making them impractical for large-scale robot learning.

More recently, generative models have shown promise for synthesizing realistic audio from visual and contextual inputs~\cite{van2016wavenet, cheng2024taming}. Advances in audiovisual learning have enabled models to generate synchronized soundtracks for silent videos \cite{li2024self, ruan2023mm}, as well as infer material properties from impact sounds \cite{zhang2017shape, clarke2022diffimpact, owens2016visually}. However, these generative approaches have primarily been explored in media synthesis and computer vision, with limited prior work applying them to robotics. Our work is the first to leverage generative audio models within a multimodal robot learning framework, demonstrating their utility in sim-to-real transfer for tasks like pouring.

\xhdr{Generative models in simulation.}
Generative models have increasingly been integrated into simulation frameworks to enhance data diversity and realism. LucidSim~\cite{yu2024learning} investigates replacing visual domain randomization with text-to-image generation models. Gen2Sim~\cite{katara2024gen2sim} generates 3D object models and their corresponding URDFs for scaling simulation assets. Eureka~\cite{ma2023eureka} uses the vast general knowledge of LLMs to generate reward functions for training policies via RL in sim. Our work departs from this trend of using generative models for augmenting existing data or capabilities. Instead, we show how they can be used for introducing entirely new sensory streams in simulation.

\vspace{-0.5em}
\section{\oursplain}
\label{sec:method}
\vspace{-0.5em}
\subsection{Framework Overview}
\label{overview}
% \vspace{-0.5em}

\begin{figure}[tp!]
\centering
\scalebox{1.0}{
    \includegraphics[width=\linewidth]{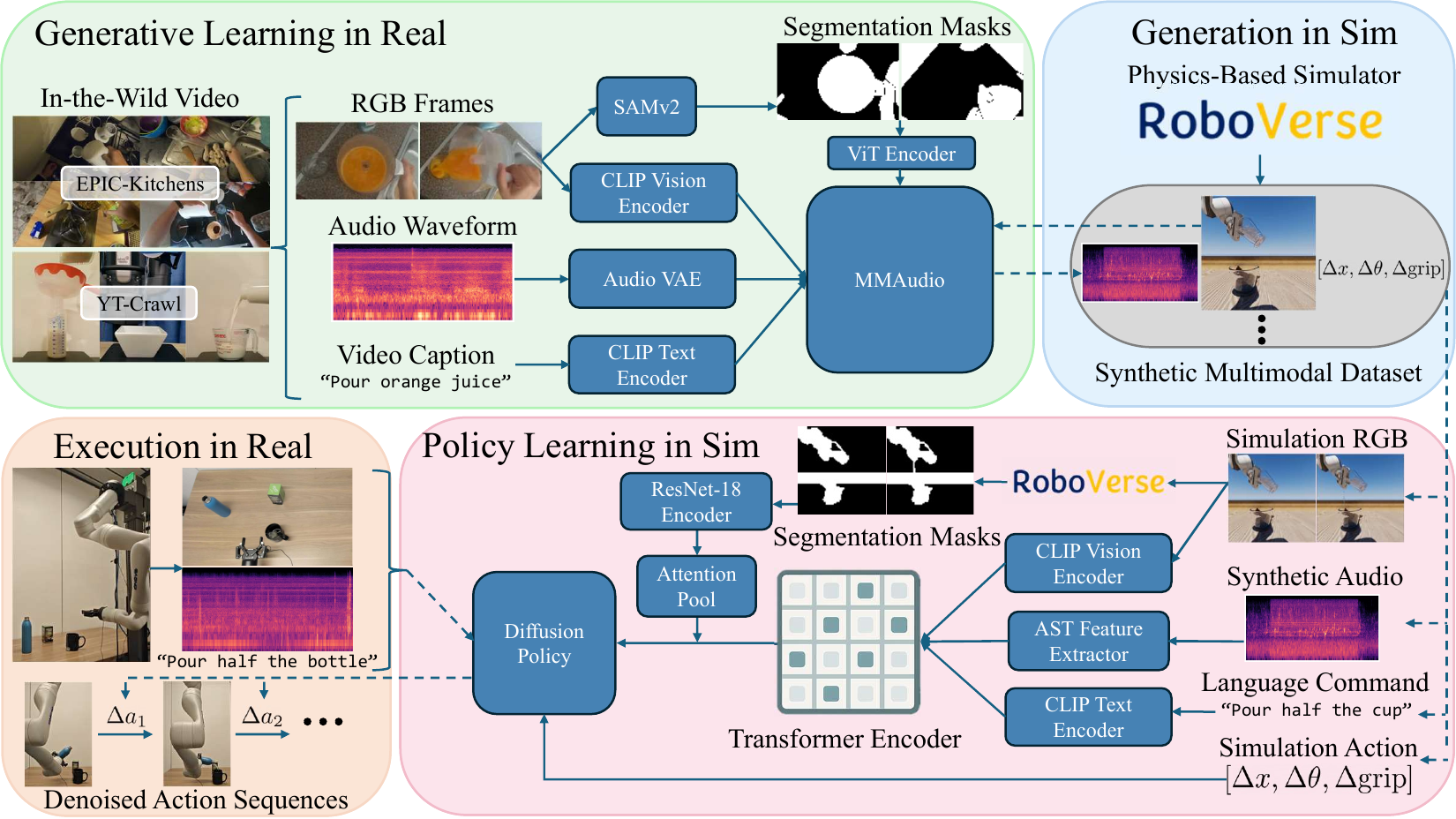}
}
\vspace{-1.5em}
\caption{\textbf{Components in the \ours instantiation for robot pouring.} We first finetune a video-to-audio diffusion model (\eg \kw{MMAudio}) on in-the-wild video. Conditioning on simulation video then enables generating multimodal simulation trajectories replete with audio. We then train a policy (\eg Diffusion Policy) on this multimodal dataset, before evaluating zero-shot on a real setup.
}
\vspace{-1.5em}
% \vspace{-0.1in}
\label{fig:method}
\end{figure}

\ours consists of two main components:

\squishlist
    \item
    1. \textbf{A physics-based simulation engine} which provides a controllable environment for robot learning. The simulator models the visual scene, rigid-body dynamics, fluid interactions, robot kinematics, \etc, and allows for the collection of visual and proprioceptive inputs.

    \item
    2. \textbf{A generative model} that conditions on simulator information and generates complementary sensory signals, such as audio, that are traditionally difficult to simulate.
\squishend

These two components interact in a hybrid pipeline, where the physics engine provides structured inputs (\eg visual renders, robot and object information), which are then used by the generative model to synthesize additional sensory signals. This interplay ensures that the generated data is physically consistent with the scene while bypassing more computationally expensive simulation processes.
We hypothesize that with sufficiently powerful components ---\ie a high-fidelity physics engine and a well-trained generative model---zero-shot transfer is achievable without finetuning on real-world data. An overview of our proposed framework is depicted in~\cref{fig:method}.

To illustrate the viability of this hypothesis, we instantiate \ours on a concrete multimodal task: robot pouring. This task inherently requires multimodal perception due to occlusion effects, visual ambiguities, and the strong dependence on auditory feedback for estimating liquid flow and container fill level.
For our simulation environment, we leverage the newly released RoboVerse framework, which provides high-quality physics simulation for robotic manipulation. For our generative model, we build upon MMAudio, a large-scale pretrained video-to-audio model, which we adapt to generate realistic pouring sounds from simulated visual inputs.

% \vspace{-1.0em}
\subsection{Generative Audio Model}
\label{subsec:audio_model}
% \vspace{-0.5em}

A primary challenge in integrating generative models for multimodal robot learning is selecting the appropriate conditioning signals. While language-conditioned audio models like AudioLDM~\cite{liu2023audioldm} can generate diverse sounds based on text prompts, they lack the fine-grained control necessary for continuous tasks like pouring. Key variables such as pour height, container volume, liquid properties, and material interactions require pixel-level information, which language alone struggles to specify.

To address this, we use MMAudio, a state-of-the-art video-to-audio model that predicts audio from raw video frames~\cite{ruan2023mm}. MMAudio employs a multimodal transformer architecture that jointly processes visual, audio, and textual features using cross-attention mechanisms. The model is trained generatively with a conditional flow matching loss. Further details are available in~\cite{ruan2023mm}.

MMAudio allows the generation of sound directly from the visual scene, capturing crucial task-relevant cues. However, we find that its zero-shot performance is inadequate for our purposes (see ablations in \cref{subsec:audio_quality}). We identify two key limitations:
% \vspace{-0.5em}
\squishlist
    \item 
    \textbf{1. MMAudio is trained on diverse but suboptimal data.} The model has been exposed to large-scale audiovisual datasets, but much of this data consists of noisy, lower-frequency, Foley sounds dramatically different from liquid sounds.

    \item
    \textbf{2. Pouring is a long-tail audio event.} The pretraining corpuses of MMAudio contain relatively few pouring instances~\cite{mei2024wavcaps, chen2020vggsound, kim2019audiocaps}. This leads to poor representations of task-specific acoustic cues, such as pitch variation as a container fills, dependencies on liquid viscosity, and transient impulse sounds at the start and stop of a pour.
\squishend

To address these limitations, we finetune MMAudio on a curated dataset of real-world pouring sounds collected from (1) EPIC-Kitchens: a large-scale first-person cooking dataset containing diverse instances of liquid manipulation~\cite{damen2020epic, damen2022rescaling}, and (2) our own YouTube crawl. Here, we extracted high-quality pouring clips from Internet video sharing platforms, featuring various liquids (\eg, water, coffee, soda, soup) and containers (\eg, glass, plastic, ceramic). Our audiovisual clips range from 5 to 60 seconds, and total 1031 videos in total. This finetuning allows the model to better capture the nuances of real-world pouring sounds, improving generalization to novel pouring conditions.

However, this adaptation introduces a new challenge: the domain gap between real-world and simulated video. Since MMAudio has been trained primarily on natural video, its performance may degrade when applied to synthetic RGB frames from simulation, which exhibit a different visual distribution. To bridge this gap, we introduce a semantic segmentation conditioning mechanism using SAMv2 (Segment Anything Model v2)~\cite{ravi2024sam}. Specifically, for each video, we extract an initial RGB frame of size $H \times W \times 3$, and obtain a segmentation mask tensor of size $H \times W \times C$, where $C$ corresponds to a predefined set of task-relevant object classes (\eg, cup, mug, water, juice). We then use SAMv2 to propagate these masks across the entire video sequence, ensuring consistent segmentation. Finally, we condition MMAudio on both the RGB frames and the segmentation masks, allowing the model to focus on semantically meaningful regions. Concretely, we introduce an additional set of projection and cross-attention parameters to inject into the transformer backbone (details in appendix).
This conditioning strategy improves robustness to the sim-to-real visual gap, and enables our model to produce more physically accurate and task-relevant audio signals.

\subsection{Simulation Framework}
\label{subsec:sim}

To effectively train multimodal robotic policies in simulation, our framework requires a simulator that meets several key desiderata. First, while works such as~\cite{pashevich2019learning} or~\cite{yu2024learning} seek to compensate for poor visual simulation via generative models, our main focus is to inject real world modalities to enable entirely new sensing capabilities. Thus, we prioritize a \textbf{photorealistic simulation environment} that provides high visual fidelity out of the box. Second, the simulator must be \textbf{computationally efficient} to support large-scale trajectory generation. Liquid simulation, in particular, is traditionally expensive due to the complexity of modeling soft-body and fluid dynamics, which often rely on particle-based solvers. Ideally, the simulator must also support parallelized data collection, enabling multiple instances to run simultaneously with efficient physics simulation and rendering. Third, the simulator should provide a \textbf{diverse and customizable environment}, both in terms of natural diversity in assets and the ability to introduce controlled variability. 

To meet these requirements, we use RoboVerse, a recently released scalable simulation platform that supports high-fidelity rendering and highly-parallelized execution. RoboVerse also offers a diverse set of predefined assets, including cups, bottles, and liquid types. To further enhance diversity, we randomly scale object dimensions (\eg cup openings, bottle heights) to diversify object geometries. RoboVerse is also equipped with a powerful DR library that allows for control over a wide-ranging set of crucial simulation parameters.

\xhdr{Environment setup.}
We begin by constructing a simulation environment that coarsely mirrors real-world conditions. We approximate camera placement based on real-world setups but maintain loose constraints (\eg precise camera extrinsics and lighting conditions are not required). Unlike traditional high-fidelity digital twin approaches, which aim for near-perfect reconstruction~\cite{jiang2021industrial}, our hybrid approach merely requires approximate modeling, as we will rely on structured domain randomization (DR) to improve generalization. Concretely, to bridge the sim-to-real gap, we apply DR to key simulation parameters, including: liquid properties (density, viscosity, color, transparency, surface friction interactions), camera pose (perturbations in position and rotation), lighting conditions (randomized intensity, shadows, and reflections), and background and table textures. A full specification for our DR setup can be found in the appendix. These variations improve generalization to unseen pouring scenes, and ensure that policies are robust to naturally occurring real-world variations.

\begin{figure}[tp!]
\centering
\scalebox{1.0}{
    \includegraphics[width=\linewidth]{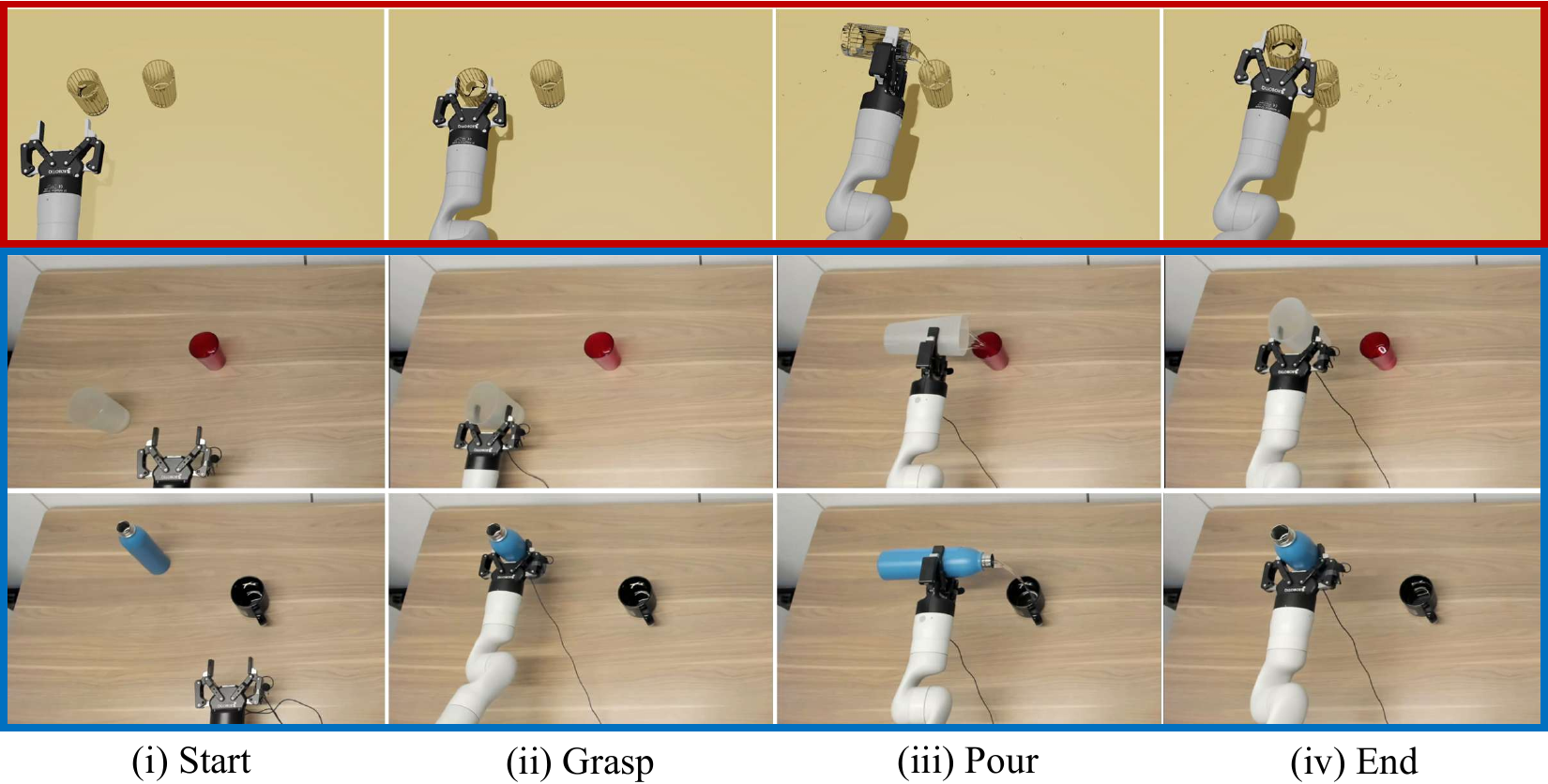}
}
\vspace{-1.5em}
\caption{\textbf{Comparison of various \textcolor{red}{simulation trajectories} and \textcolor{blue}{real-world executions}.} Trajectories motion-planned in our photorealistic simulator mirror execution traces in the real world.}
\vspace{-1.5em}
% \vspace{-0.1in}
\label{fig:sim_vs_real}
\end{figure}

\xhdr{Generating vision and action.}
To generate diverse and structured pouring trajectories, we use a motion planning-based pouring strategy. Every trajectory consists of a pouring container and a receiving container, with the robot initialized from the same home position. The pouring container is randomly pre-filled to a $70\% \pm 5\%$ capacity. We sample a random grasp point within a 5\% offset from the container center, ensuring a consistent but slightly varied grasp strategy. Then, the robot moves the container above the receiving container, ensuring alignment with its opening geometry. This pour height is again randomly sampled within predefined limits to ensure variability. Next, one of four discrete pouring amounts is sampled (one-quarter, one-half, three-quarters, or full pour). Based on the initial fill level and the selected pouring fraction, we compute a deterministic tilt angle to achieve the desired liquid transfer. Finally, once pouring is completed, the robot untilts the container, and the episode terminates. We show in~\cref{fig:sim_vs_real} these various keypoints in both trajectories generated in simulation, as well as real-world execution traces from learned policies.
For simplicity, we use motion planning to generate trajectories in simulation, but our framework could be adapted to use any number of approaches depending on the chosen task (\eg reinforcement learning or other pretrained policies with verification mechanisms, or virtual teleoperation for collecting human-guided demonstrations.)

\xhdr{Generating audio.}
To generate synchronized multimodal data, we run a parallelized simulation pipeline with two threads. The first thread executes the motion-planned pouring trajectory, recording RGB frames, proprioception, and action sequences. However, the RGB frames from cameras corresponding to the real-world setup may be heavily occluded depending on the pouring scenario, or misaligned with the camera poses of our finetuning dataset. Thus, we crucially also collect video from a dedicated virtual camera positioned frontal to the receiving container. This view is in distribution with the real-world pouring videos used for finetuning our video-to-audio generative model. The second thread then leverages this virtual stream to process completed trajectories and generate audio. 

\section{Experiments}
\label{sec:experiments}
\vspace{-0.5em}
In this section, we begin by describing our policy learning and evaluation methodology, as well as our real-world setup. We then dive into results which definitively answer two main questions: (1) First, to what extent do policies learned in simulation via \ours transfer to the real world? (2) Second, how well does the generated audio from \ours align with real-world pouring sounds, both perceptually and in terms of task-relevant metrics? 

\subsection{Evaluation Methodology: Baselines and Benchmarks}
\label{subsec:eval_setup}

We choose a state-of-the-art policy learning approach to train and evaluate our framework. Namely, we learn a diffusion policy~\cite{chi2023diffusion, liu2024maniwav}, a denoising diffusion model which conditionally generates actions given state, vision and audio inputs. Importantly, our policy is trained from scratch using only proprioception, vision, and audio data generated entirely by \ours, as described in~\cref{subsec:sim}. To systematically analyze the impact of multimodality, we also train a variant which omits audio. 

For evaluation, we establish a benchmark spanning diverse pouring conditions. Our evaluation considers variations in container materials (\eg plastic, paper, metal), liquid types (\eg water, juice, soda, hot liquids), and occlusion levels (\eg transparent vs. opaque containers). By introducing realistic distribution shifts, we assess the adaptability and robustness of policies trained in simulation. The full suite of conditions can be seen in~\cref{tab:main_eval}, with details available in the appendix. For each condition, we choose three pairs of random starting positions for the pouring and receiving containers.
For each position pair, we deploy the policy four times, once with each of the language instructions described in~\cref{subsec:sim}, leading to twelve total evaluations for each condition. 
All policies are evaluated zero-shot, meaning \emph{no demonstrations or real robot data have been employed.}

To compensate for differences in relative difficulty or target volume across language commands and container setups, our evaluation metric is Normalized Mean Absolute Error (NMAE):

\vspace{-1.8em}
\begin{center}
\begin{equation}
\label{eqn:nmae}
NMAE = \frac{\mid\text{Actual Poured Amount - Desired Target Amount}\mid}{\text{Desired Target Amount}}.
\end{equation}
\end{center}
\vspace{-0.5em}

Note that the desired target amount is given by the known initial volume in the pouring container and the language command (\eg pour half). We average over the three distinct locations per command, before averaging across the four commands.

\subsection{Physical Setup and Implementation Details}
\label{subsec:physical_setup}

\xhdr{Robot hardware and setup.}
Our experiments are conducted on a Kinova Gen3 robot equipped with a Robotiq 2F-85 adaptive gripper. Monocular RGB is provided via a single Logitech BRIO 4K web camera mounted in an egocentric position. Audio is obtained via a MAONO omnidirectional USB lapel microphone mounted at the end effector, capturing 24-bit audio at 192 kHz. For all tasks, we generate 6-DoF Cartesian space delta end-effector commands at a policy frequency of 10 Hz. The MoveIt IK library converts these commands to a desired 7-DoF joint action. A full depiction of our workspace and data setup can be found in the appendix. 

\xhdr{Policy architecture details.} For our audiovisual diffusion policy, we follow~\cite{liu2024maniwav} and use a CLIP-pretrained ViT-B/16 model~\cite{dosovitskiy2020image} to encode RGB frames. 
We use an audio spectrogram transformer (AST)~\cite{gong2021ast} to encode audio. Audio is downsampled from 24-bit, 192 kHz into 16-bit, 16 kHz, before conversion to a log-mel spectrogram via FFT with temporal window 400, hop length 160 and 64 mel filterbanks. 
Complete training hyperparameters are available in the appendix.

\subsection{Results: Zero-Shot Sim-to-Real Transfer}
\label{subsec:zero_shot}

Our main results are presented in \cref{tab:main_eval}, demonstrating strong sim-to-real transfer. In particular, policies trained with \ours achieve high success rates in real-world pouring tasks, suggesting that our simulation pipeline produces transferable policies without requiring real-world data for training (average NMAE: \textbf{0.46}). Critically, the multimodal vision + audio diffusion policy variant outperforms the vision-only baseline, highlighting the importance of auditory feedback in our pouring task. Specifically, including audio induces a \textbf{23.3\%} average reduction in NMAE. Pouring to and from opaque containers tends to benefit the most of audio, with an average NMAE reduction of \textbf{29.4\%} (compared to a \textbf{16.2\%} reduction for transparent containers). This is intuitive: in settings where visual feedback is limited, audio plays a crucial role in estimating flow rate and container fill levels. 

\begin{table*}[t!]
\centering
\setlength{\tabcolsep}{4pt}
\begin{tabular}{lcccc}
\toprule
Opaque & \begin{tabular}[c]{@{}c@{}}water-\\ white cup-\\ red cup\end{tabular} & \begin{tabular}[c]{@{}c@{}}coffee-\\ metal thermos-\\ paper cup\end{tabular} & \begin{tabular}[c]{@{}c@{}}sake-\\ sake carafe-\\ sake cup\end{tabular} & \begin{tabular}[c]{@{}c@{}}water-\\ metal thermos-\\ metal mug\end{tabular}  \\ 
\midrule
\diffp \kw{(V)} & $0.54 \pm 0.23$ & $0.54 \pm 0.20$ & $0.68 \pm 0.30$ & $0.47 \pm 0.29$ \\
\diffp \kw{(V + A)} & \textbf{0.44 $\pm$ 0.19} & \textbf{0.33 $\pm$ 0.17} & \textbf{0.43 $\pm$ 0.28} & \textbf{0.37 $\pm$ 0.15} \\ 
\midrule
Transparent & \begin{tabular}[c]{@{}c@{}}water-\\ white cup-\\ plastic cup\end{tabular} & \begin{tabular}[c]{@{}c@{}}juice-\\ plastic bottle-\\ plastic cup\end{tabular} & \begin{tabular}[c]{@{}c@{}}soda-\\ metal can-\\ plastic cup\end{tabular} & \begin{tabular}[c]{@{}c@{}}juice-\\ plastic bottle-\\ glass mug\end{tabular} \\ 
\midrule
\diffp \kw{(V)} & $0.53 \pm 0.20$ & $0.46 \pm 0.20$ & $0.43 \pm 0.20$ & $0.49 \pm 0.21$ \\
\diffp \kw{(V + A)} & \textbf{0.42 $\pm$ 0.27} & \textbf{0.38 $\pm$ 0.21} & \textbf{0.39 $\pm$ 0.22} & \textbf{0.43 $\pm$ 0.18} \\ 
\bottomrule
\end{tabular}
\caption{\textbf{Main evaluation results on our pouring benchmark (lower is better).} We report average normalized mean absolute error (NMAE) and 1 std. dev. error. The top four tasks all involve opaque containers, and the bottom four tasks involve translucent containers. Results are computed over twelve seeds (four language commands evaluated for three random locations each.)
}
\label{tab:main_eval}
\vspace{-.5em}
\vspace{-1em}
\end{table*}

\subsection{Results: Assessing Audio Generation Quality}
\label{subsec:audio_quality}

\begin{figure}[tp!]
\centering
\scalebox{1.0}{
    \includegraphics[width=\linewidth]{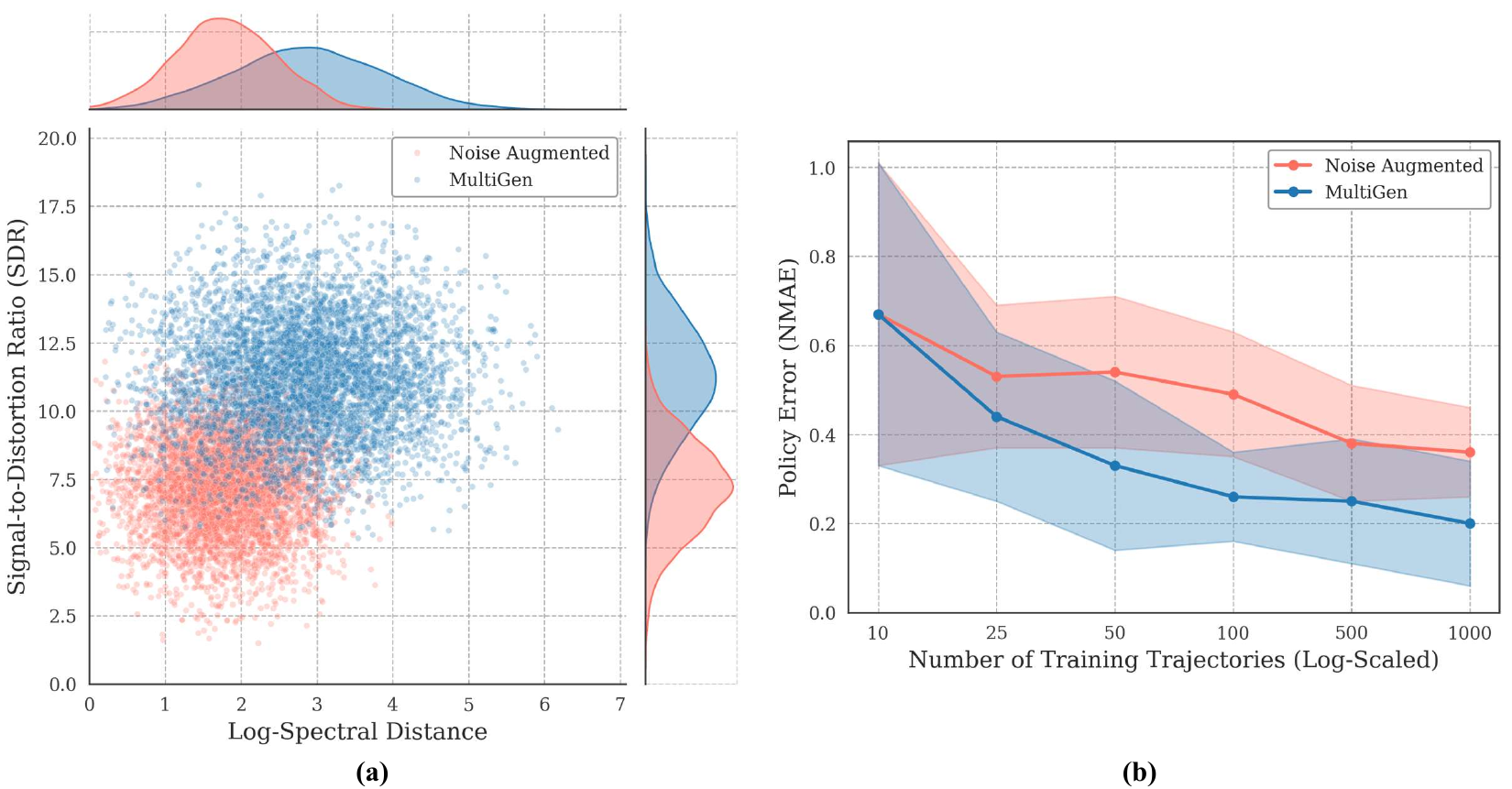}
}
\vspace{-1.5em}
\caption{\textbf{Comparison between \ours and standard data augmentation used in robotics.} \textbf{(a) Left:} \ours produces audio that is both more diverse (higher Log-Spectral Distance) and more accurate (higher Signal-to-Distortion Ratio) than traditional additive noise augmentation. \textbf{(b) Right:} This higher quality audio allows policies trained with \ours to demonstrate better scaling properties (\ie lower policy error across increasing dataset sizes.)
}
\vspace{-1.5em}
% \vspace{-0.1in}
\label{fig:ablation}
\end{figure}

In this section, we rigorously evaluate the impact of integrating a generative audio model into the simulation pipeline. Our goal is to quantify the benefits of \ours by comparing it to explicit data augmentation techniques and measuring how well the generated audio supports policy learning. We evaluate across three key axes:

\squishlist
    \item
    1. \textbf{Diversity:} How diverse is the synthetic audio data generated by \ours?

    \item
    2. \textbf{Fidelity:} How well does \ours audio match real-world pouring sounds?

    \item
    3. \textbf{Usefulness:} Can \ours enable scalable learning of multimodal policies?
\squishend

To establish a baseline, we collect 10 pouring demonstrations in the real-world, using a random selection of pouring and receiving containers. Pouring heights and initial container locations are randomized (details available in the appendix.) Then, we follow the data augmentation protocol from \kw{ManiWAV}~\cite{liu2024maniwav}, where background noises are overlaid onto training audio. Specifically, \cite{liu2024maniwav} augments audio from human demonstrations with i) randomly sampled environment noises from ESC-50~\cite{piczak2015esc} and ii) robot motor sounds from 10 task-specific trajectories. These are overlayed onto the original audio signal with 50\% probability. We generate 1000 samples using this augmentation approach, and 1000 samples from our finetuned generative audio model. For the latter, we condition on RGB and SAMv2 masks from the 10 pouring demonstrations to generate the corresponding audio.

To assess diversity, we compute the \emph{minimum} log-spectral distance (LSD) between each generated sample and the collected demonstrations. This allows us to measure how much variability each approach introduces relative to real-world audio. To assess fidelity, we measure the signal-to-distortion ratio (SDR). We then plot these two quantities against each other, as shown in~\cref{fig:ablation}a.  We observe that \ours generates a significantly broader range of audio samples, producing a wider marginal distribution over spectral distances compared to explicit augmentations. Crucially, despite greater diversity, \ours samples remain physically consistent, exhibiting higher SDR than augmented audio, even for samples far from the training distribution. This suggests that \ours achieves strong generalization, synthesizing diverse yet dynamically accurate audio.

Finally, to validate the impact on policy learning, we train a diffusion policy using either dataset. We compare both policies on a subset of the benchmark described in~\cref{subsec:eval_setup}, including two opaque and two transparent pour settings. As shown in~\cref{fig:ablation}b, results indicate that \ours-trained policies significantly outperform augmentation-based policies. Moreover, \ours-trained policies exhibit stronger scaling with increasing number of generated trajectories, compared to \naive augmentation. Taken together, these ablations confirm that \ours generates more realistic, task-relevant audio, and that this property directly leads to improved multimodal policy learning.

\vspace{-0.75em}
\section{Conclusion}
\label{sec:conclusion}
\vspace{-0.75em}

In this work, we introduced \ours, a novel framework for integrating generative multimodal simulation into robot learning. By augmenting physics-based simulators with large-scale generative models, we demonstrated that sim-to-real policy learning can leverage rich sensory feedback beyond vision and proprioception. We instantiated \ours in the context of robot pouring, a task where auditory cues are critical.
\ours enabled zero-shot sim-to-real transfer, facilitating generalization to challenging real-world pouring without requiring real robot sensory or action data.
Our results highlight the broader potential of generative models to fundamentally expand simulation capabilities by injecting entirely new sensory modalities into training environments. 

\section{Limitations}
While \ours demonstrates strong multimodal sim-to-real transfer for robot pouring, a few limitations remain. First, our approach relies on projecting real-world conditions into simulation, requiring approximate alignment between simulated and real environments. Although we only enforce loose constraints (\eg approximate camera placement and correct robot embodiment), this still necessitates some manual setup. However, ongoing advancements in real-to-sim techniques can further automate this process, making it more scalable and reducing human effort. Second, our trained policies exhibit limited generalization to extreme container geometries. For instance, significantly larger or irregularly shaped containers can introduce unmodeled physical dynamics (e.g., unexpected liquid sloshing) that degrade performance. Expanding the diversity of simulated training assets and incorporating adaptive policies that reason about novel container shapes could mitigate this issue. Third, while our policies successfully execute pouring tasks in controlled real-world settings, they do not explicitly account for collision avoidance when navigating cluttered environments. In practical deployments, additional obstacle-aware motion planning or integrated perception modules would be required to prevent unintended collisions with surrounding objects.

\clearpage
% The acknowledgments are automatically included only in the final and preprint versions of the paper.
\acknowledgments{
The authors would like to thank Qiyang Li and Phillip Isola for helpful discussions on experimental design. Philipp Wu also contributed significantly to an earlier version of this work. Yuvan Sharma assisted with real-world setup. RW is supported in part by the Toyota Research Institute and ONR MURI. PA holds concurrent appointments as a Professor at UC Berkeley and as an Amazon Scholar. This paper describes work performed at UC Berkeley and is not associated with Amazon.
}

\bibliography{refs}  % .bib

\newpage
\appendix
\vspace{-0.5em}
\section{Details on Evaluation Benchmarks}
\label{app:eval_details}

In this section, we detail the physical setups involved in our main benchmark (\ie the evaluations in~\cref{tab:main_eval}), and our ablation (\ie the 10 base pouring demonstrations used in~\cref{subsec:audio_quality}, and the evaluation setting for~\cref{fig:ablation}b.)

\subsection{Main Benchmark}

\xhdr{Water-White Cup-Red Cup.} The pouring container is a tall opaque plastic cup with total volume 491 mL. The receiving container is a plastic red cup with total volume 284 mL. The pouring liquid is tap water.

\xhdr{Coffee-Metal Thermos-Paper Cup.} The pouring container is a tall, thin blue metal thermos with total volume 500 mL. The receiving container is a disposable paper cup with total volume 287 mL. The pouring liquid is hot instant coffee.

\xhdr{Sake-Sake Carafe-Sake Cup.} The pouring container is an irregularly shaped sake carafe with total volume 216 mL. The receiving container is a small sake cup with total volume 35 mL. The pouring liquid is room temperature Junmai sake.

\xhdr{Water-Metal Thermos-Metal Mug.} The pouring container is a tall, thin blue metal thermos with total volume 500 mL. The receiving container is a black metal mug with total volume 300 mL. The pouring liquid is hot water.

\xhdr{Water-White Cup-Plastic Cup.} The pouring container is a tall opaque plastic cup with total volume 491 mL. The receiving container is a clear disposable plastic cup with total volume 356 mL. The pouring liquid is tap water.

\xhdr{Juice-Plastic Bottle-Plastic Cup.} The pouring container is a SimplyOrange bottle with total volume 450 mL. The receiving container is a clear disposable plastic cup with total volume 356 mL. The pouring liquid is orange juice with pulp.

\xhdr{Soda-Metal Can-Plastic Cup.} The pouring container is asmall round Coca Cola soda can with total volume 222 mL. The receiving container is a clear disposable plastic cup with total volume 356 mL. The pouring liquid is Coca Cola soda.

\xhdr{Juice-Plastic Bottle-Glass Mug.} The pouring container is a SimplyOrange bottle with total volume 450 mL. The receiving container is a black glass mug with total volume 364 mL. The pouring liquid is orange juice with pulp.

\subsection{Ablation Benchmark}
\label{app:abl_benchmark}

To collect 10 pouring demonstrations, we choose the following subset from the previously described 8 main evaluation settings. 

\squishlist
    \item 
    \textbullet\quad water-white cup-red cup (x2)

    \item 
    \textbullet\quad water-metal thermos-metal mug (x2)

    \item 
    \textbullet\quad water-white cup-plastic cup (x2)

    \item 
    \textbullet\quad juice-plastic bottle-plastic cup (x2)

    \item 
    \textbullet\quad juice-plastic bottle-glass mug (x2)
    
\squishend

For each setting, we randomly choose a starting location for the pouring and receiving containers, which are both distinct from the locations used in the main evaluation benchmark. We also randomly choose one of the 4 different pour commands (quarter, half, three-quarters, all). Demonstrations are collected via Apple Vision Pro teleoperation~\cite{park2024using}.

For evaluation, we choose 2 transparent container and 2 opaque container settings, namely:

\squishlist
    \item 
    1. water-white cup-plastic cup

    \item 
    2. juice-plastic bottle-plastic cup

    \item 
    3. water-white cup-red cup

    \item 
    4. water-metal thermos-metal mug
    
\squishend

Similar to our main benchmark, we evaluate 3 random positions per setting, across 4 distinct pouring commands (quarter, half, three-quarters, all). Note that the evaluation setting is completely within distribution with respect to the original 10 pouring demonstrations.
\vspace{-0.5em}
\section{Generative Model Details}
\label{app:mmaudio_details}

In this section, we provide additional details on the MMAudio video-to-audio generative model, including architectural modifications and finetuning design choices.

\subsection{MMAudio Architecture}
\label{app:mmaudio_architecture}

The MMAudio model follows a modular, multimodal architecture for conditional video-to-audio generation. Concretely, video, audio, and text streams are encoded separately via modality-specific encoders and fused in a series of multimodal transformer blocks to generate audio latents autoregressively. For further details on the original MMAudio design, please refer to~\citet{cheng2024taming}.

\paragraph{Segmentation Pathway.} We introduce a new input modality during finetuning: semantic segmentation masks aligned with each RGB frame. For each mask of shape $H \times W \times C$ (where $C$ is the number of semantic classes), we apply a lightweight convolutional encoder to extract spatial features:
\begin{itemize}
    \item Two 3$\times$3 convolution layers with ReLU activation,
    \item Global average pooling to obtain a per-frame embedding $s_t \in \mathbb{R}^{d_\text{seg}}$ (we use $d_\text{seg} = 512$),
    \item A projection layer to map $s_t$ to the transformer embedding dimension (1024).
\end{itemize}

\paragraph{Injection Strategy.} We inject the projected segmentation embedding $s_t$ into the model as an additional global conditioning vector alongside the existing visual and text-based conditioning. Specifically, $s_t$ is concatenated with the pooled visual and text tokens and added to the conditioning stream used by all multimodal transformer blocks. This avoids modifying the original CLIP encoder or the SyncFormer path, preserving the pretrained backbone while allowing the model to leverage segmentation information during finetuning.

This injection is implemented using a separate projection MLP and attention-based fusion mechanism, where the transformer dynamically integrates $s_t$ alongside the timestep embedding and existing global context. By gating the contribution of $s_t$ (via learned scalars), the model can modulate its reliance on segmentation during finetuning.

For ease of understanding this injection implementation, we provide the PyTorch pseudo-code below:

\begin{lstlisting}[language=Python, caption={Segmentation mask fusion module. A learnable gate modulates the contribution of segmentation embeddings relative to the original global conditioning vector.}, label={lst:seg_fusion}]
class SegmentationFusion(nn.Module):
    def __init__(self, dim):
        super().__init__()
        self.gate = nn.Linear(dim * 2, 1)

    def forward(self, c_g, seg_embed):
        """
        c_g: (B, T, D) global conditioning vector, see Cheng et al.
        seg_embed: (B, T, D) segmentation embedding
        returns: (B, T, D) fused conditioning vector
        """
        x = torch.cat([c_g, seg_embed], dim=-1)     # (B, T, 2D)
        alpha = torch.sigmoid(self.gate(x))         # (B, T, 1)
        return alpha * seg_embed + (1 - alpha) * c_g
\end{lstlisting}

\subsection{Ablation: Zero-Shot vs. Finetuned}

One natural hypothesis is that the pretrained MMAudio model is sufficient to generate high-quality pouring sounds out-of-the-box. To test this, we conducted a controlled ablation comparing the \emph{zero-shot} performance of the pretrained model against the \emph{finetuned} variant described in~\cref{subsec:audio_model}. Our goal was to measure the extent to which finetuning improves the fidelity, diversity, and downstream utility of the generated audio.

\xhdr{Finetuning Setup.} Our finetuning dataset and input modalities are described in~\cref{subsec:audio_model}. Our finetuning architecture is described in~\cref{app:mmaudio_architecture}. We froze the CLIP vision encoder and updated only the following components:
\begin{itemize}
    \item The multimodal transformer backbone (attention and projection layers),
    \item The audio decoder (diffusion UNet),
    \item The segmentation cross-attention module (newly added),
    \item The final linear projection layers for spectrogram generation.
\end{itemize}

Audio was represented as log-mel spectrograms using a 64-bin mel filterbank, 16 kHz sampling rate, 25ms window, and 10ms hop size. During training, the spectrogram was denoised from Gaussian noise over 100 diffusion steps. For optimization, we used AdamW with a learning rate of 1e-5, and a cosine decaying scheduler with 500-step warmup. The batch size was 32, and the losses involved the default MMAudio losses (L1 waveform loss + spectral convergence loss + perceptual loss using AudioCLIP embeddings). The diffusion sampler was DDIM with 50 diffusion timesteps. We finetuned for a total of 100K steps ($\sim$40 epochs).

\xhdr{Evaluation Setup.} We use a subset of 100 simulation trajectories as the basis for our ablation. For each video, we generated synthetic audio using both the zero-shot and finetuned models, conditioned on RGB only for the former, and RGB + segmentation masks for the latter.

\xhdr{Quality Metrics.} We evaluated generated audio across three axes:

\begin{itemize}
    \item \textbf{Fidelity}: Measured via Signal-to-Distortion Ratio (SDR), computed between the generated waveform and ground-truth microphone audio. Higher is better.
    \item \textbf{Spectral Accuracy}: Assessed using Log-Spectral Distance (LSD) between generated and real spectrograms. Lower indicates better spectral alignment.
    \item \textbf{Diversity}: We compute the minimum LSD between each generated sample and the training set to assess the marginal spread of synthetic samples. This tests whether the model produces varied outputs or mode-collapses to common templates.
\end{itemize}

\xhdr{Utility Metrics.} We further evaluated the practical utility of each model by training the same multimodal diffusion policy architecture on the generated trajectories from each variant. We then evaluated both policies in the real world on a 4-condition benchmark (2 opaque, 2 transparent pours) using the protocol described in~\cref{app:abl_benchmark}. Performance was measured using Normalized Mean Absolute Error (NMAE) of the poured volume.

\xhdr{Quantitative Results.} The results are summarized below:

\begin{center}
\begin{tabular}{lccc}
\toprule
\textbf{Metric} & \textbf{Zero-Shot} & \textbf{Finetuned} & \textbf{Relative Change} \\
\midrule
Signal-to-Distortion Ratio (SDR) $\uparrow$ & 7.8 dB & \textbf{11.3 dB} & +45\% \\
Log-Spectral Distance (LSD) $\downarrow$ & 1.95 & \textbf{1.43} & -26.7\% \\
Audio Diversity (min-LSD to train set) $\uparrow$ & 0.67 & \textbf{1.02} & +52.2\% \\
Policy NMAE $\downarrow$ & 0.50 & \textbf{0.37} & -26.0\% \\
\bottomrule
\end{tabular}
\end{center}

\xhdr{Qualitative Observations.} Audio generated by the zero-shot model often exhibited unnatural characteristics—overly low-frequency hums, repetitive bubbling artifacts, or missing transient features such as pour onset and cutoff. These artifacts were substantially reduced in the finetuned model, which produced audio with realistic variations in pitch, volume, and dynamics that aligned with the visual pouring cues. Spectrogram inspection revealed tighter alignment of temporal events and improved high-frequency detail in the finetuned outputs.

\xhdr{Conclusion.} These results conclusively show that finetuning is essential for adapting pretrained generative audio models to domain-specific robotic tasks. Despite the general capabilities of MMAudio, its pretrained version lacks the task-specific inductive biases and visual grounding needed for physically plausible pouring sounds. Finetuning on a small, curated dataset enables the model to capture subtle acoustic dynamics critical for multimodal policy learning and sim-to-real transfer.

\vspace{-0.5em}
\section{Additional Training Details}
\label{app:train_details}

This section contains additional training details to supplement~\cref{subsec:physical_setup}. Our RGB observations are $224 \times 224 \times 3$ frames, with a frame stack history of 2 timesteps. Our audio observations are 533 samples per timestep, with a horizon of 120 timesteps. Finally, our proprioception observations are 7-DoF joint angles, with a history of 2 timesteps. 

Our diffusion policy predicts a 10-dim action vector (6D rotation representation) with a horizon of 16 timesteps. We use a DDIM sampler with 50 diffusion timesteps. Our scheduler is the Glide cosine scheduler~\cite{nichol2021glide}, with beta start and end at 0.0001 and 0.02, respectively. We train with a batch size of 64, using the AdamW optimizer and a learning rate of 0.0001. Learning rate is cosine decayed after 500 warmup steps, with a total of 500 training epochs (250 steps per epoch). We incorporate weight decay of 1e-6 on all trainable parameters.
\vspace{-0.5em}

\end{document}